# Large Language Models are Highly Aligned with Human Ratings of Emotional Stimuli

**Mattson Ogg (mattson.ogg@jhuapl.edu)**
Research and Exploratory Development Department
Johns Hopkins University Applied Physics Laboratory
11100 Johns Hopkins Road, Laurel, MD 20723

**Chace Ashcraft (chace.ashcraft@jhuapl.edu)**
Research and Exploratory Development Department
Johns Hopkins University Applied Physics Laboratory
11100 Johns Hopkins Road, Laurel, MD 20723

**Ritwik Bose (rik.bose@jhuapl.edu)**
Research and Exploratory Development Department
Johns Hopkins University Applied Physics Laboratory
11100 Johns Hopkins Road, Laurel, MD 20723

**Raphael Norman-Tenazas (raphael.norman-tenazas@jhuapl.edu)**
Research and Exploratory Development Department
Johns Hopkins University Applied Physics Laboratory
11100 Johns Hopkins Road, Laurel, MD 20723

**Michael Wolmetz (michael.wolmetz@jhuapl.edu)**
Research and Exploratory Development Department
Johns Hopkins University Applied Physics Laboratory
11100 Johns Hopkins Road, Laurel, MD 20723

## Abstract

**Emotions exert an immense influence over human behavior and cognition in both commonplace and high-stress tasks. Discussions of whether or how to integrate large language models (LLMs) into everyday life (e.g., acting as proxies for, or interacting with, human agents), should be informed by an understanding of how these tools evaluate emotionally loaded stimuli or situations. A model's alignment with human behavior in these cases can inform the effectiveness of LLMs for certain roles or interactions. To help build this understanding, we elicited ratings from multiple popular LLMs for datasets of words and images that were previously rated for their emotional content by humans. We found that when performing the same rating tasks, GPT-4o responded very similarly to human participants across modalities, stimuli and most rating scales (r = 0.9 or higher in many cases). However, arousal ratings were less well aligned between human and LLM raters, while happiness ratings were most highly aligned. Overall LLMs aligned better within a five-category (happiness, anger, sadness, fear, disgust) emotion framework than within a two-dimensional (arousal and valence) organization. Finally, LLM ratings were substantially more homogenous than human ratings. Together these results begin to describe how LLM agents interpret emotional stimuli and highlight similarities and differences among biological and artificial intelligence in key behavioral domains.**

## Introduction

Emotion is essential to biological intelligence, informing almost every aspect of human behavior and everyday life (Pessoa, 2008). The centrality of emotion to the human experience is underscored by its stature as a major focus of psychological research for over a century (Wundt, 1912).

The growing body of literature studying the cognitive capacities of artificial intelligence systems (Abdurahman et al., 2024; Ivanova, 2025; Jones & Bergen, 2025; Ogg et al., 2025) would be well served by providing an analogous understanding of how artificial intelligence conceptualizes, interprets or reacts to emotional stimuli. This will be critical for numerous applications whether for interacting with emotional human agents (e.g., as assistants) or taking on roles that human agents might otherwise perform that require a high level of emotional intelligence (e.g., as a therapist or teacher).

Among the many debates still unresolved in the psychological literature is one regarding the appropriate way to organize emotional responses and their physiological correlates. Does human emotion exist within five discreet categories (happiness, anger, sadness, fear, disgust; Ekman, 1992; Izard, 1992), or does it exist along two broader dimensions (arousal and valence; Russell, 1980)? The popularity of large language models and their increasing ubiquity invite inquiries into how similar the emotional processing of these models is to human emotion processing. Moreover, can we clarify whether artificial intelligence uses frameworks analogous to those within human psychology to understand emotion?

## Methods

We obtained multiple public data sets of images (OASIS, Kurdi, Lozano, and Banaji, 2017; NAPS, Marchewka Żurawski, Jednoróg, and Grabowska, 2014) and words (ANEW, Bradley & Lang, 1999; Stevenson, Mikels, and James, 2007) that were previously rated by large cohorts of humans for their emotional qualities (all rated for arousal and valence, ANEW was also rated for each of the five emotion categories). We then elicited ratings from several popular high performing LLMs (GPT-4o, GPT-4o-mini, Gemma2-9B Llama3-8B and Solar 10.7B) using paradigms that replicated each original study. Exact prompts for each trial varied to match each study but were generally of the following form: "Please rate the word respectful for Happiness: 1=not at all, to 5=extremely. Respond only with a number. Please use the full range of the scale to make your responses rather than relying on only a few points." We ran twenty LLM "participants" (e.g., an LLM initialized for a complete run of trials) for each rating paradigm and dataset (at a temperature of 1.0). Any items that activated a model's content filter were removed from further analysis for that model. We then compared each LLM cohort with the human cohorts aggregated at the item-level (mean and standard deviation across participants for each emotion rating scale for each item).

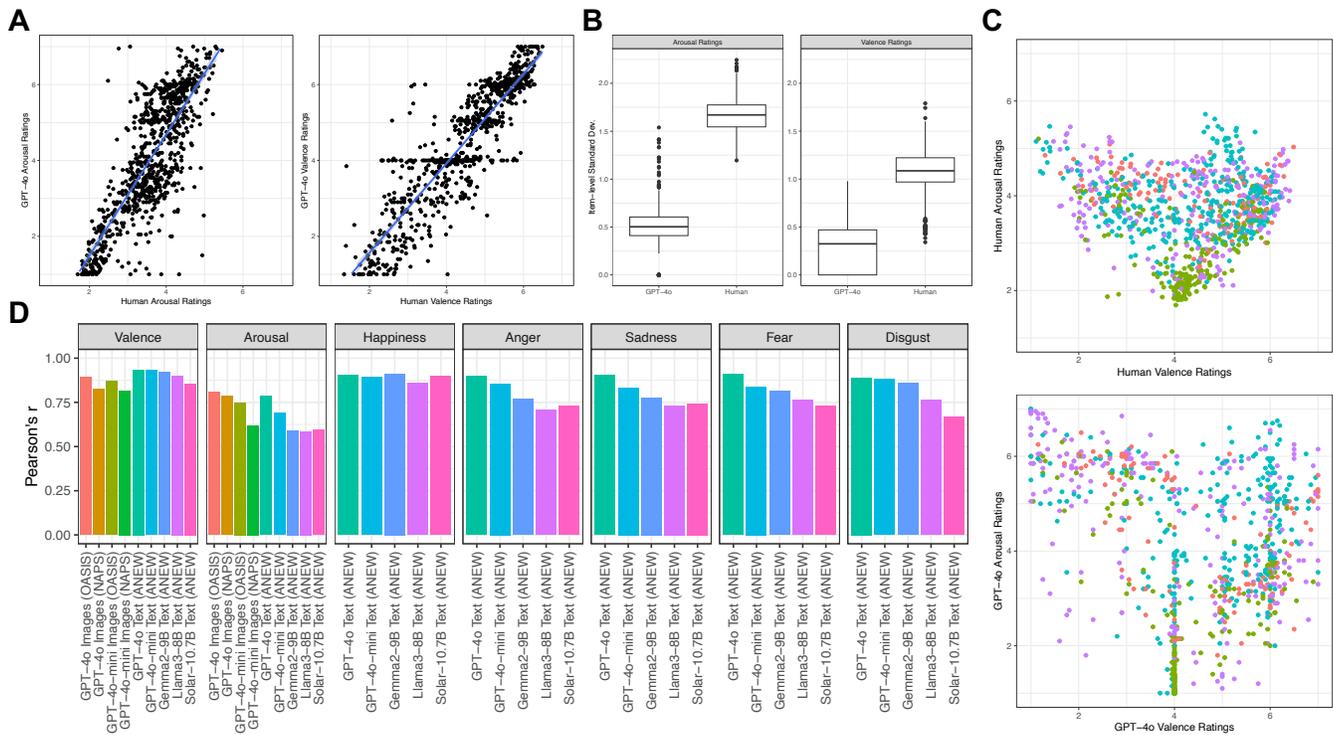

*Figure 1 Summary of Human and LLM Behavioral Ratings of Emotional Stimuli.* A to C: Example comparison of GPT-4o and human ratings of images in the OASIS dataset for arousal and valence. GPT-4o ratings of arousal ($r = 0.81$) and valence ($r = 0.89$, both $p < 0.001$) were highly correlated with human ratings (A, blue line indicates linear fit), and produced a broadly similar V-shaped distribution (C). However, GPT-4o participant responses were more homogenous compared to human ratings (B, Wilcoxon rank sum test comparing the standard deviation of responses across items, $W > 660$, $p < 0.001$). Similar results were obtained for the GPT-4o models across datasets and emotion rating scales (D, Summary of all LLM-Human comparisons for ratings of text, ANEW, and image, OASIS and NAPS, datasets, all $p < 0.001$).

## Results and Conclusion

Large language models, and GPT-4o in particular, were well aligned with human ratings for different emotion scales (all Pearson correlation $p < 0.001$). Figure 1 summarizes the results for the OASIS experiments for GPT-4o (Figure 1A through 1C) along with a summary of all models across datasets (Figure 1D). GPT-4o generally produced the highest alignment with human ratings of both text and images for all of the five-category emotion scales ($r = 0.89$ to $0.93$), higher than arousal and valence ($r = 0.79$ and $0.90$, respectively) for the ANEW dataset. Overall arousal ratings were least aligned between humans and LLMs ($r = 0.59$ to $0.81$), while happiness ratings were best aligned ($r = 0.86$ to $0.91$). Note that in all cases, the standard deviation of ratings produced by the LLM participants for each item was lower than for human participants (all Wilcoxon rank sum tests, $p < 0.001$).

These results illuminate the surprisingly close correspondence between human and LLM raters for emotional stimuli. This suggests that biological and artificial intelligence are well aligned on a critical dimension of behavior, one that influences cognition and mediates interactions between agents. While our findings show stronger LLM-human alignment within a five-category framework than a two-dimensional model, further research is needed to determine whether this pattern reflects inherent properties of emotion representation or artifacts of language model training.

## Acknowledgments

We acknowledge support from the Independent Research and Development (IRAD) Fund from the Research and Exploratory Development Mission Area of the Johns Hopkins Applied Physics Laboratory. Feedback from GPT-4o was used to assist in the editing this report.

## References

Abdurahman, S., Atari, M., Karimi-Malekabadi, F., Xue, M. J., Trager, J., Park, P. S., Golazizian, P., Omrani, A., & Dehghani, M. (2024). Perils and opportunities in using large language models in psychological research. *PNAS Nexus*, *3*(7), pgae245. https://doi.org/10.1093/pnasnexus/pgae245

Bradley, M. M., & Lang, P. J. (1999). *Affective Norms for English Words (ANEW): Instruction Manual and Affective Ratings*. *31*(1), 25–36.

Ekman, P. (1992). An argument for basic emotions. *Cognition and Emotion*, *6*(3–4), 169–200. https://doi.org/10.1080/02699939208411068

Ivanova, A. A. (2025). How to evaluate the cognitive abilities of LLMs. *Nature Human Behaviour*, *9*(2), 230–233. https://doi.org/10.1038/s41562-024-02096-z

Izard, C. E. (1992). Basic emotions, relations among emotions, and emotion-cognition relations. *Psychological Review*, *99*(3), 561–565. https://doi.org/10.1037/0033-295x.99.3.561

Jones, C. R., & Bergen, B. K. (2025). *Large Language Models Pass the Turing Test* (No. arXiv:2503.23674). arXiv. https://doi.org/10.48550/arXiv.2503.23674

Kurdi, B., Lozano, S., & Banaji, M. R. (2017). Introducing the Open Affective Standardized Image Set (OASIS). *Behavior Research Methods*, *49*(2), 457–470. https://doi.org/10.3758/s13428-016-0715-3

Marchewka, A., Żurawski, Ł., Jednoróg, K., & Grabowska, A. (2014). The Nencki Affective Picture System (NAPS): Introduction to a novel, standardized, wide-range, high-quality, realistic picture database. *Behavior Research Methods*, *46*(2), 596–610. https://doi.org/10.3758/s13428-013-0379-1

Ogg, M., Bose, R., Scharf, J., Ratto, C., & Wolmetz, M. (2025). *Turing Representational Similarity Analysis (RSA): A Flexible Method for Measuring Alignment Between Human and Artificial Intelligence* (No. arXiv:2412.00577). arXiv. https://doi.org/10.48550/arXiv.2412.00577

Pessoa, L. (2008). On the relationship between emotion and cognition. *Nature Reviews Neuroscience*, *9*(2), 148–158. https://doi.org/10.1038/nrn2317

Russell, J. A. (1980). A circumplex model of affect. *Journal of Personality and Social Psychology*, *39*(6), 1161–1178. https://doi.org/10.1037/h0077714

Stevenson, R. A., Mikels, J. A., & James, T. W. (2007). Characterization of the Affective Norms for English Words by discrete emotional categories. *Behavior Research Methods*, *39*(4), 1020–1024. https://doi.org/10.3758/BF03192999

Wundt, W. M. (1912). *An Introduction to Psychology,*. G. Allen, Limited.